\documentclass[letterpaper]{article} 
\usepackage{aaai23}  
\usepackage{times}  
\usepackage{helvet}  
\usepackage{courier}  
\usepackage[hyphens]{url}  
\usepackage{graphicx} 
\usepackage{natbib}  
\usepackage{caption} 
\usepackage{newfloat}
\usepackage{listings}
\usepackage{amsmath,amssymb}
\usepackage{xcolor}
\usepackage[colorlinks=true, citecolor=citecolor, linkcolor=linkcolor]{hyperref}
\definecolor{citecolor}{HTML}{0071BC}
\definecolor{linkcolor}{HTML}{ED1C24}
\usepackage{multirow}
\usepackage{tabulary}
\usepackage{booktabs}
\usepackage{xspace}
\usepackage{soul}
\usepackage{array}
\usepackage[ruled, linesnumbered, vlined]{algorithm2e}
\usepackage{subcaption}
\usepackage{floatrow}
\newfloatcommand{capbtabbox}{table}[][\FBwidth]

\newcommand{\pdata}{p_{\rm{data}}}

\newcommand{\diff}{{\rm{d}}}

\def\rvh{{\mathbf{h}}}

\def\rvx{{\mathbf{x}}}

\def\rvz{{\mathbf{z}}}

\def\calB{{\mathcal{B}}}
\def\calD{{\mathcal{D}}}

\def\calN{{\mathcal{N}}}
\def\calT{{\mathcal{T}}}
\def\calX{{\mathcal{X}}}
\def\calZ{{\mathcal{Z}}}

\newlength\savewidth\newcommand\shline{\noalign{\global\savewidth\arrayrulewidth
		\global\arrayrulewidth 1pt}\hline\noalign{\global\arrayrulewidth\savewidth}}
\newcolumntype{x}[1]{>{\centering\arraybackslash}p{#1pt}}
\newcommand{\tablestyle}[2]{\setlength{\tabcolsep}{#1}\renewcommand{\arraystretch}{#2}\centering\small}
\newcommand{\tabincell}[2]{\begin{tabular}{@{}#1@{}}#2\end{tabular}}

\makeatletter
\DeclareRobustCommand\onedot{\futurelet\@let@token\@onedot}
\def\@onedot{\ifx\@let@token.\else.\null\fi\xspace}

\def\eg{\emph{e.g}\onedot} 
\def\ie{\emph{i.e}\onedot} 
 
\def\etc{\emph{etc}\onedot} 
 
\def\etal{\emph{et al}\onedot}
\makeatother

\title{Local Manifold Augmentation for Multiview Semantic Consistency}
\author{
	Yu Yang\equalcontrib, Wing Yin Cheung\equalcontrib, Chang Liu, Xiangyang Ji
}
\affiliations{

Tsinghua University, BNRist\\
\texttt{yang-yu16@foxmail.com, zhangyx20@mails.tsinghua.edu.cn,\\ 
\{liuchang2022, xyji\}@tsinghua.edu.cn}
}

\begin{document}
\nocopyright
\maketitle

\begin{abstract}
Multiview self-supervised representation learning roots in exploring semantic consistency across data of complex intra-class variation. 
Such variation is not directly accessible and therefore simulated by data augmentations.
However, commonly adopted augmentations are handcrafted and limited to simple geometrical and color changes, which are unable to cover the abundant intra-class variation. 
In this paper, we propose to extract the underlying data variation from datasets and construct a novel augmentation operator, named local manifold augmentation (LMA). 
LMA is achieved by training an instance-conditioned generator to fit the distribution on the local manifold of data and sampling multiview data using it.
LMA shows the ability to create an infinite number of data views, preserve semantics, and simulate complicated variations in object pose, viewpoint, lighting condition, background \etc. 
Experiments show that with LMA integrated, self-supervised learning methods such as MoCov2 and SimSiam gain consistent improvement on prevalent benchmarks including CIFAR10, CIFAR100, STL10, ImageNet100, and ImageNet. 
Furthermore, LMA leads to representations that obtain more significant invariance to the viewpoint, object pose, and illumination changes and stronger robustness to various real distribution shifts reflected by ImageNet-V2, ImageNet-R, ImageNet Sketch \etc.
\end{abstract}

\section{Introduction}

\begin{figure}[t!]
    \centering
    \begin{subfigure}[b]{0.99\linewidth}
        \centering
        \includegraphics[width=\textwidth]{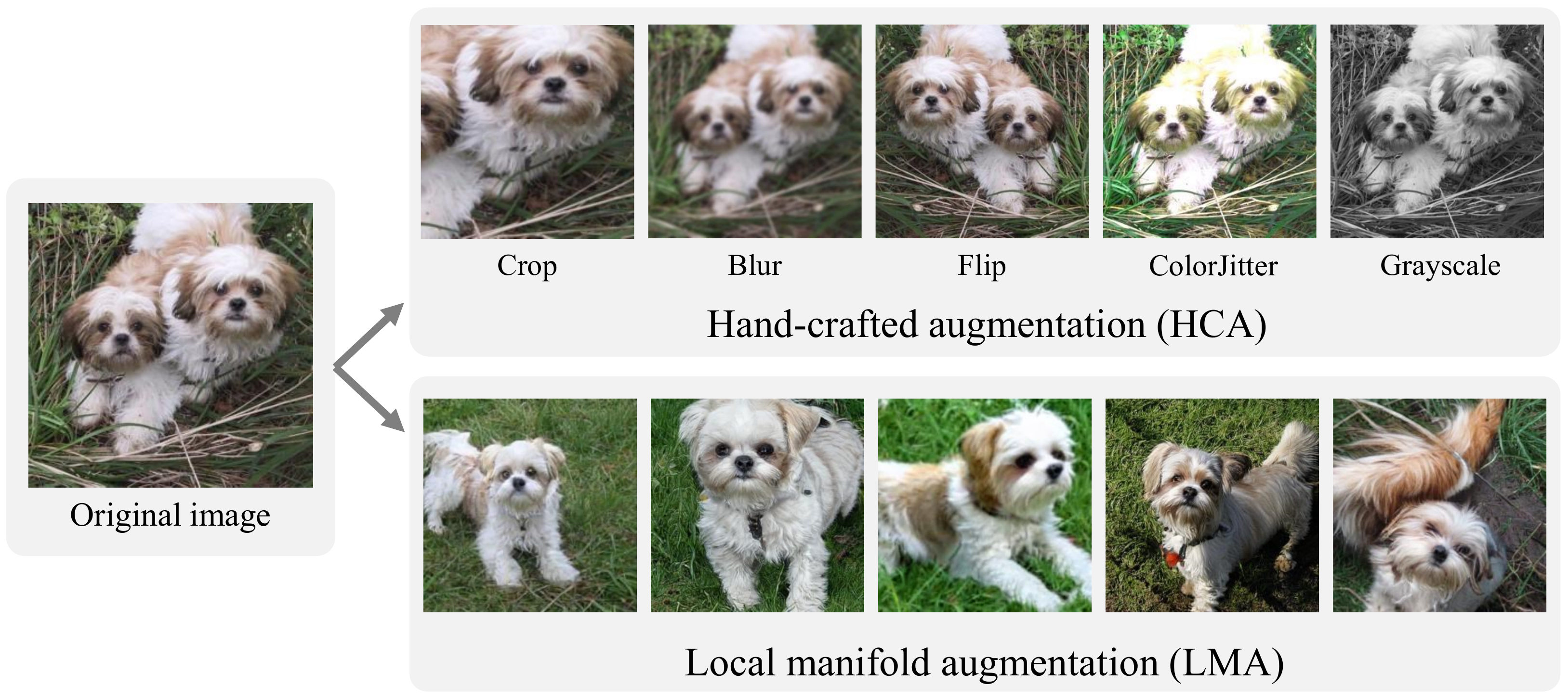}
        \caption{Visual effect}
        \label{fig:teaser_effect}
    \end{subfigure}
    \begin{subfigure}[b]{0.99\linewidth}
        \centering
        \includegraphics[width=\textwidth]{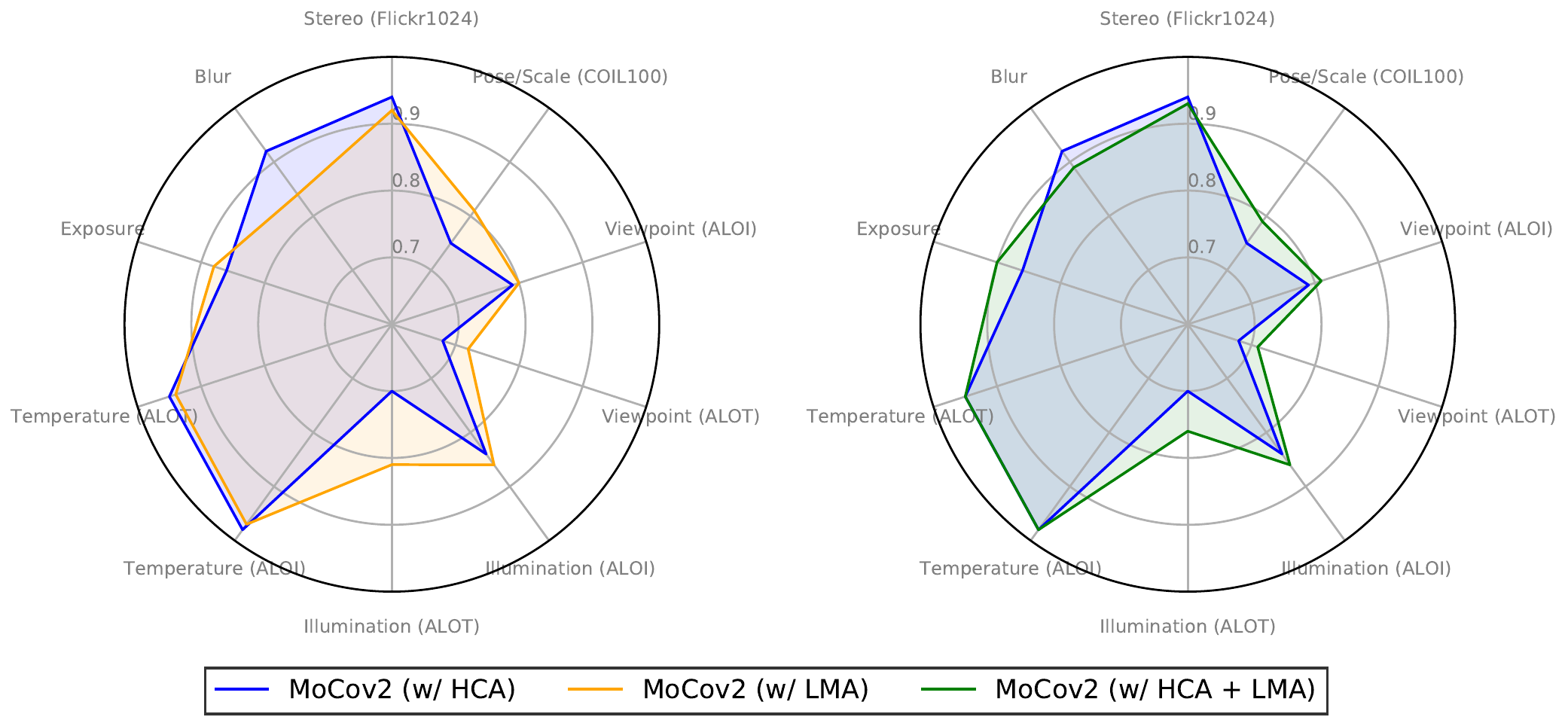}
        \caption{Representation invariance}
        \label{fig:teaser_invariance}
    \end{subfigure}
    \caption{Comparison of handcrafted augmenetation (HCA) and local manifold augmentation (LMA). 
    (a) Visual effect comparison: while HCA leads to simple geometrical and appearance change, LMA consequences more complicated changes, \eg object pose, viewpoint, lighting condition, background \etc. 
    (b) LMA helps MoCov2 to obtain representations that are more  invariant to changes in object pose, viewpoint, and illumination.}
    \label{fig:teaser}
\end{figure}

With a vision of leveraging massive data for effective visual representation learning, there has been a surging interest in self-supervised learning (SSL). As a prevalent SSL paradigm, multiview self-supervised learning are driven by the goal of learning semantic consistency across data with intra-class variation, which is usually achieved by attracting positive pair of views~\cite{byol2020grill,simsiam2021chen,barlow2021zbontar,vicreg2021bardes} or meanwhile repelling negative pair of views~\cite{cpc2018oord,simclr2020chen,cmc2020tian,moco2020he,swav2020mathilde}. 
Thanks to the learned invariance to nuisance variation~\cite{ericsson2021self}, multiview SSL have shown to yield representations that strongly generalizes to different domains with significant distribution shifts~\cite{mitrovic2021representation} and various downstream tasks~\cite{liu2020self,xie2021propagate,xiao2021posecontrast}.


Therefore, one crucial point in multiview SSL is to access the multiview data with nuisance variation.
Such variations are not directly available and therefore are usually simulated with data augmentation techniques such as cropping image patches~\cite{cpc2018oord} and changing the color~\cite{cmc2020tian}.
To this end, effective augmentation strategies have been constructed through a dedicated composition of elementary transformations to provide abundant multiview data and have shown to greatly improve the SSL performance~\cite{simclr2020chen,mocov2_2020chen}. 
Despite so, such a strategy is still limited by the handcrafted operators which are difficult to design and can hardly simulate complicated nuisance variations such as varied object pose, viewpoint, lighting condition, and background.

Although such complicated variation is challenging to be simulated by handcrafted transformation, it is ubiquitous in the collected datasets.
It is quite common that a dataset contains multiple images that depict the same scene or object with varied nuisance factors.
These images naturally serve as a valuable source of multiview data with complicated geometrical or appearance changes failed by handcrafted operators but still critical for representation learning. 

In this paper, we study extracting the underlying data variation from datasets and construct an augmentation operator named local manifold augmentation (LMA).
Straightforward ways to achieve this goal could be regarding the $k$ nearest neighbors ($k$NN) in the dataset as nuisance views and 
traversing the vicinity in latent space mapped to the whole dataset by generative models (\eg GAN).
However, these methods either suffer from limited views or face the challenge to preserve the semantics of augmented views~\cite{jahanian2021generative}.
Instead, we model distribution on the local manifold of data with an instance-conditioned GAN (IC-GAN)~\cite{icgan2021casanova} and repurpose it for augmentation. 
In particular, $k$ nearest neighbors of an image in the dataset are first identified with $k$NN algorithm, and a generator is conditioned on given images and trained to generate images that have a similar appearance to the conditioned images.
To this end, the trained generator learns to transform images to their neighbors that vary in some nuisance factors.  
Since the modeling approach takes advantage of both $k$NN and generative models, LMA instantiated with IC-GAN can create infinite views yet preserve semantics.


Although the multiview data can be significantly enriched by LMA,
direct integration of LMA into existing multiview SSL hurts the performance in practice.
We analyze that LMA has the potential to reduce the overall quality of training data due to the notorious mode collapse issue of GANs. 
To mitigate this issue, we apply LMA with a certain probability when introduced into SSL, which ends up with training representation networks on a mixture of real and generated data and preserves the diversity of training data.

Fig.~\ref{fig:teaser} compares the handcrafted augmentation (HCA) and LMA.
In contrast to HCA which only leads to simple changes such as crop and color distortion, LMA can consequence more complicated changes in object pose, lighting condition, viewpoint, background \etc (Fig.~\ref{fig:teaser_effect}). 
We further test the integration of LMA into MoCov2~\cite{mocov2_2020chen} that requires negative pairs, SimSiam~\cite{simsiam2021chen} that does not.
Comprehensive experiments on prevalent benchmarks including CIFAR10, CIFAR100, STL10, ImageNet100, and ImageNet show that LMA consistently improves MoCov2 and SimSiam. 
Furthermore, quantitative evaluations show that the LMA helps to improve the representation invariance to changes in object poses, viewpoints, and illumination (Fig.~\ref{fig:teaser_invariance})
and strengthen the representation robustness to various distribution shifts in ImageNet-V2, ImageNet-R, ImageNet Sketch \etc.

Our contribution can be summarized as: 
\begin{itemize}
    \item [(1)] A novel data augmentation method, named local manifold augmentation (LMA), which can provide more complicated data variation for SSL. 
    \item [(2)] A method that integrates LMA into SSL algorithms, which is empirically shown to consistently improve the performance on prevalent benchmarks such as CIFAR10, CIFAR100, STL10, ImageNet100, and ImageNet, and gain more invariant and robust representations.
\end{itemize}

\section{Related Work}

\subsection{Self-supervised Representation Learning}
A rich body of methods devise pretext tasks~\cite{agrawal2015learning,doersch2015unsupervised,zhang2017split,wang2017transitive,wang2015unsupervised,pathak2016context,pathak2017learning,misra2016shuffle,mahendran2018cross,larsson2016learning,kim2018learning,jenni2018self} where labels come from data itself.
In the recent trend of SSL, contrastive learning~\cite{cpc2018oord,cmc2020tian,simclr2020chen,moco2020he} becomes one of the most popular approaches and show its great power in learning transferable representations that even outperform supervised one in various transfer learning tasks~\cite{moco2020he,mocov2_2020chen}.
Contrastive learning draws presentations of positive data pairs together and push apart representations of negative pairs, where the positive pairs are usually obtained by applying data augmentation to create two different views of the same data point.
More recently, contrasting to negative samples are further proven to be unnecessary by non-contrastive methods~\cite{byol2020grill,simsiam2021chen,barlow2021zbontar,vicreg2021bardes}.
In both contrastive and non-contrastive learning methods, invariance with respect to particular transformations, also known as data augmentation, is still a critical motive for SSL algorithms.

\subsection{Data Augmentation}
Data augmentation is a crucial technique for creating multi-view data in SSL.
Low-level image processing operators are employed as data augmentation tools.
For example, CPC~\cite{cpc2018oord} employs image crops for multi-view data and CMC~\cite{cmc2020tian} takes different color channels of different color-space images as multi-view data.
SimCLR~\cite{simclr2020chen} first integrates multiple data augmentation including crop, resize, color distortion, Gaussian blurring \etc and empirically show its importance in contrastive learning.
The data augmentation pipeline is tuned later which further helps to improve the performance~\cite{mocov2_2020chen}.
More recently, the data augmentation pipeline is further enriched with multi-crop~\cite{swav2020mathilde} and background removal~\cite{tomasev2022pushing}, which significantly contribute to the improvement of performance.
In our work, we also study improving self-supervised representation learning methods with novel data augmentation techniques. 
In contrast to widely-used ones that are mostly low-level visual transformation, our work attempts to introduce high-level visual transformation that can provide richer multi-view data sources.
The most related work to ours is Jahanian \etal~\cite{jahanian2021generative} that creates multi-view data by traversing GAN latent space. 
Our work differs from it in that we employ a particular GAN for creating multi-view data.

\subsubsection{Steerable GANs} 
Besides its stunning capability of synthesizing high-fidelity images, the steerable generation of GANs has also attracted a lot of research interests. 
Earlier works show that smooth transition between images can be obtained with interpolation in the latent space~\cite{gan2014goodfellow} and a mixture of ingredients from two images can be obtained by mixing the latent variables~\cite{stylegan2019karras,stylegan22020karras}. 
More recent works are more interested in finding ways to manipulate generation to consequence expected changes. 
One line of works discover the semantic direction in GAN latent space guided by pre-trained attribute classifiers~\cite{abdal2021styleflow,bau2018gan,nitzan2020face,patashnik2021styleclip,shen2020interfacegan,wu2021stylespace,yang2021semantic}.
However, these works requires external supervision to discover meaningful directions and therefore do not suite the self-supervised learning setting. 
Another line of works~\cite{gld2020voynov,harkonen2020ganspace,shen2021closed,spingarn2020gan,ramesh2018spectral,zhu2021low,esser2020disentangling,choi2021not} search latent directions without external human supervision.
However, it is not guaranteed that the discovered latent manipulation is benificial to self-supervised representation learning and a recent study shows that random traversal is good enough~\cite{jahanian2021generative}.
Unlike the above works that traverse latent space of GANs, we explore the generation of one particular GAN, \ie IC-GAN~\cite{icgan2021casanova}, to create data augmentation for SSL learning.

\section{Method}

\subsection{Multiview Modeling}
An implicit visual concept $c$, \eg a scene or an object, is presented to computers in form of images that probably vary as different views $v(c)$.
Multiview representation learning pulls together multiview representations to learn representations $f(c)$ that are invariant to nuisance views, \ie $f(v(c)) = f(c)$. 
One challenge is that multiview data is not always available. 
Exsiting methods address this issue by transforming a given data point to obtain different views,
\begin{equation}
    v(c) = t(i(c)),~t \sim \calT,
\end{equation}
where $i(c)$ denotes the an image that represent $c$, which is available from the collected dataset, $t$ denotes a particular data transformation sampled from a distribution of transformation $\calT$.
However, since the transformation is usually constructed by composing a variety of hand-crafted operators, it is difficult to precisely depict the underlying multiview distribution $P(v|c)$.

In this paper, we consider leveraging distribution of local data manifold to sample multiview data.
We are motivated by that the collected dataset contains redundant data that reflect nuisance variation of the same visual concept.
Therefore, we first acquire a local data manifold distribution $Q(\rvx|i(c))$, where $\rvx$ denotes a data point, and regard it as an approximation of the underlying multiview distribution, \ie $Q(\rvx|c) \approx P(v|c)$.
The multiview data is then obtained by sampling data from the approximated multiview distribution,
\begin{equation}
    v(c) \sim Q(\rvx|i(c)).
\end{equation}

\subsection{Multiview Data from Local Data Manifold}
In this section, we start from analyzing two simple methods, $k$NN and traversing GAN latent space, for modeling local data distribution.
After that, we introduce our method that repurposes IC-GAN~\cite{icgan2021casanova} to achieve this goal.
Finally, the merits of our method are discussed.

\subsubsection{Sampling from $k$NN} 
Given a specific similarity metric, the local data manifold at a data point can be approximated with its $k$ nearest neighbors.
Formally, given a pair of data $\rvx_i, \rvx_j$, the similarity is measured with distance induced by certain embedding function $f_\phi:\calX\rightarrow\mathbb{R}^{d}$, \ie $\lVert f_\phi(\rvx_i) - f_\phi(\rvx_j) \rVert_2$.
This embedding function $h$ can be a convolutional neural network learned with SSL methods.  
Then sampling multiview data from the distribution of local data manifold can be written as sampling from its nearest neighbor set, 
\begin{equation}\label{eq:knn_t}
    \rvx_{t,i} \sim k\text{NN}(\rvx_i),
\end{equation}
where $k\text{NN}(\rvx_i)$ denotes the set of $k$ nearest neighbors of $\rvx_i$ in $\calD$ based on the similarity metric and we use $\sim$ to denote a uniformly random sample from a set.
However, since the nearest neighbor set is a finite set containing limited views, views created by this method is highly restricted.
Hence, $k$NN may fail to provide sufficient data variation for multiview representation learning.

\subsubsection{Traversing latent space} 
Supposing the global data distribution is available, one can achieve multiview data sampling by traversing the data distribution.
In particular, the data distribution can be modeled with deep generative models such as GANs.
GANs implicitly approximate the target data distribution by building up a model to favor a sampling process.
This sampling process is realized by first sampling a latent variable from prior distribution, and then transforming the latent variable with into data with generator network.
From the distribution perspective, generator network is like mapping a latent distribution, usually a noise distribution, to the target distribution.
Traversing data distribution is therefore feasible by traversing the latent space.
 
Formally, given a trained generator $G:\calZ\rightarrow\calX$ mapping a random latent variable $\rvz_i\in\calZ$ to an image $\rvx_i = G(\rvz_i)\in\calX$.
Another view of $\rvx_i$ can be sampled by first sampling a latent perturbation $\epsilon\sim p(\epsilon)$ and then forwarding the perturbed $\rvz_i$ to generator
\begin{equation}\label{eq:trav_gan}
    \rvx_{t,i} = G(\rvz_i + \epsilon).
\end{equation} 
Nonetheless, this method faces two challenges.
First, this method can only generates multiview data for generated data instead of real data, which significantly limit its application.
Second, such traversal is chanllenging to be controlled without external supervision. 
It is difficult to avoid trivial changes that can be hardly perceived and excessive alteration that intensively alter the semantic concept of the image. 
These issues makes traversing GAN latent space contribute little to improving self-supervised representation learning performance~\cite{jahanian2021generative}.

\subsubsection{Sampling from local data distribution} 
To allieviate the above issues, we consider directly modeling the local data distribution. 
This modeling is also known as instance-conditioned generative model, which has been explored by recently proposed instance-conditioned GAN (IC-GAN)~\cite{icgan2021casanova} and shown great power in image synthesis.
Instead of focusing on synthesis, we concern its ability to model local data distribution and repurpose IC-GAN for sampling multiview data. 

\begin{figure}[t]
    \centering
    \includegraphics[width=\linewidth]{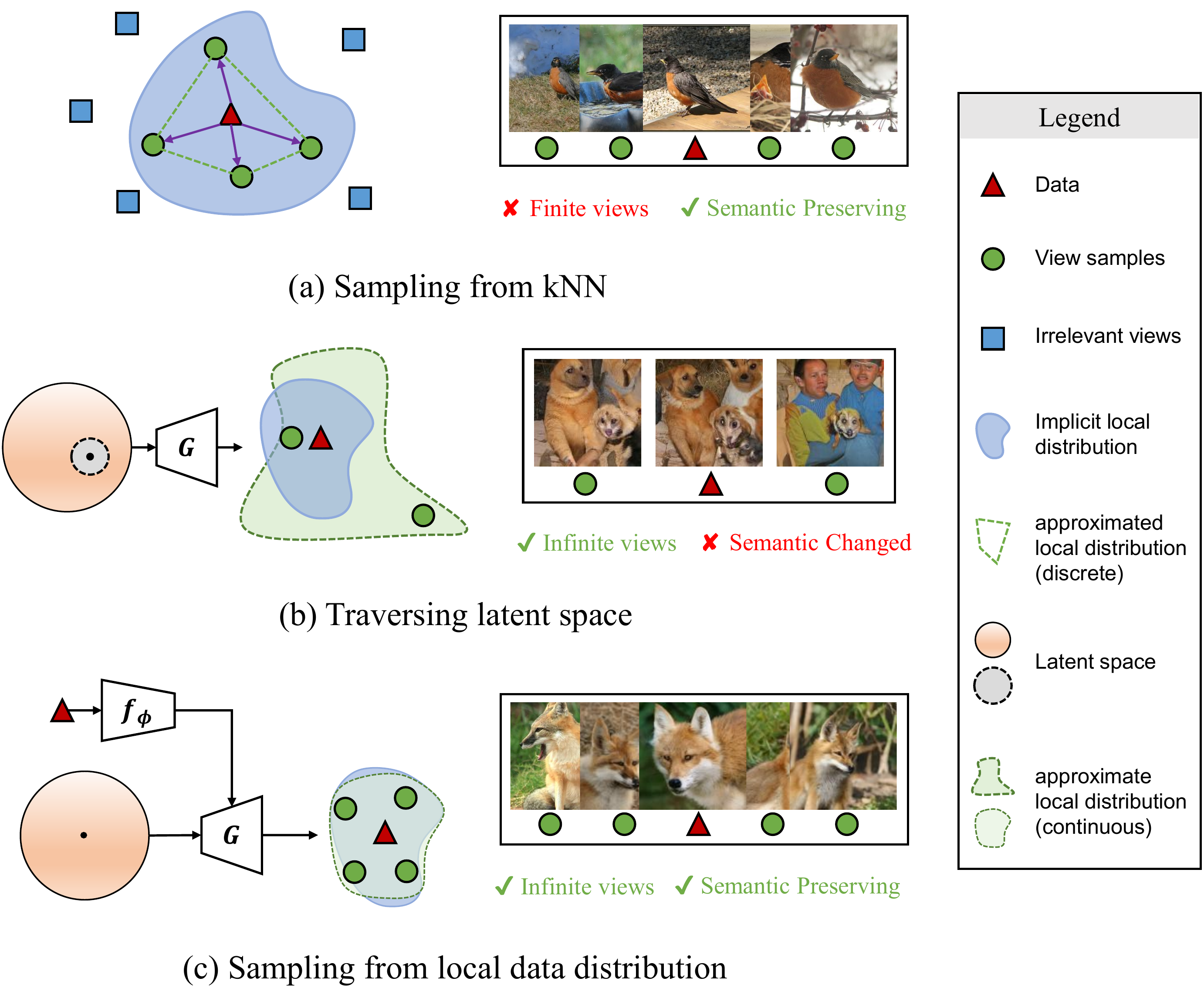}
    \caption{\textbf{Illustration} of different instantiation of local manifold augmentation.}
    \label{fig:comparison}
\end{figure}

Concretely, IC-GAN~\cite{icgan2021casanova} decomposes the real data distribution into a mixture of conditional distributions and task a conditional generator to fit each conditional data distribution.
Formally, the real data distribution is approximated as $\pdata=\int p(\rvx | \rvx_i)p(\rvx_i)\diff\rvh\approx\frac{1}{N}\sum_i p(\rvx|\rvx_i)$, where $p(\rvx|\rvx_i)$ represents a local data distribution at a data sample from dataset $\rvx_i\sim\calD$. 
Given this decomposition, a conditional generator $G:\calZ\times\{\rvh_i\}_{i=1}^{M}\rightarrow\calX$ is constructed to take as input a random variable $\rvz\in\calZ$ and as condition a embedding vector $f_\phi(\rvx_i)$ and outputs an image
\begin{equation}
    \rvx = G(\rvz, f_\phi(\rvx_i)),
    ~~\rvz\sim p_{\rm{z}},
    ~\rvx_i\sim\calD,
\end{equation}
where $p_{\rm{z}}$ denotes the prior distribution of $\rvz$, typically a normal distribution.
$G$ is trained against a discriminator to tell if generated images are realistic nearest neighbors of $\rvx_i$.
In this way, $G$ implicitly models the distribution the target local data distribution $p(\rvx|\rvx_i)$ with its ability of sampling from local data distribution. 



We wrap the generation process of IC-GAN, including the feature extraction of conditioning image, as a data augmentation operation, denoted as $\text{LMA}(\cdot)$. 
In particular, given an image $\rvx_i$, its augmented view $\rvx_{t,i}$ is created through
\begin{equation}\label{eq:gda}
    \rvx_{t,i} = \text{LMA}(\rvx_i) = G(\rvz, f_\phi(\rvx_i)),
    ~~\rvz\sim \calN(0, 1).
\end{equation}

\subsubsection{Discussion} As illustrated in Fig.~\ref{fig:comparison}, $k$NN can generate non-trivial views but only supports finite number of views.
On the contrary, traversing data distribution can provide inifinite number of views but is challenging to avoid trivial and excessive views.
Traversing local data distribution can be understood as integrating these two methods and therefore is able to create inifinite number of appropriate views. 

\subsection{Learning Discriminative Semantic Invariance}\label{sec:ssl_LMA}

\begin{algorithm}[t]
	\footnotesize
	\DontPrintSemicolon
	\SetAlgoLined
	\SetKwInOut{Input}{Inputs}
	\SetKwInOut{Output}{Output}
	\Input{
		\par
		\begin{tabular}{ll}
            $\calD$, $\calT$ &~~~~dataset and hand-crafted augmentation \\
			\text{LMA}&~~~~local manifold augmentation \\
            $\alpha\in(0,1)$  &~~~~probability of applying \text{LMA} \\
            $f_\theta$  &~~~~representation extractor initialized with $\theta$ \\
	    \end{tabular}
    }
	\For{$t\gets1$ \KwTo $T$}{ 
        $\calB \gets \left\{\rvx_i\sim\calD\right\}_{i=1}^B$ \tcp*{{sample a mini-batch from dataset}}
        \For{$i\gets1$ \KwTo $B$}{
            \tcp{generative data augmentation}
            with probability $\alpha$ apply $\rvx_{i,t_1}\gets\text{LMA}(\rvx_i)$

            with probability $\alpha$ apply $\rvx_{i,t_2}\gets\text{LMA}(\rvx_i)$

            \tcp{normal data augmentation}
            $\rvx_{i,t_1}\gets t_1(\rvx_{i,t_1})$, where $t_1\sim\calT$
            
            $\rvx_{i,t_2}\gets t_2(\rvx_{i,t_2})$, where $t_2\sim\calT$
        }
        $\theta\gets\texttt{ssl-alg}(f_\theta, \{(\rvx_{i,t_1}, \rvx_{i,t_2})\}_{i=1}^{B})$
        \tcp*{update $\theta$ with SSL algorithm}
    }
\Output{representation network $f_\theta$}
\caption{Self-supervised learning with \text{LMA}}
\label{alg:gda_ssl}
\end{algorithm}

We integrate \text{LMA} into exsiting multiview representation learning approaches as in Algorithm~\ref{alg:gda_ssl}. 
Note that \text{LMA} should be applied prior to other data augmentation $\calT$. 
The reason is that feature extractor used in IC-GAN is pre-trained to be invariant to these data augmentation. 
\text{LMA} therefore would erase the effect of other data augmentation. 

It is noteworthy that \text{LMA} is applied to each data point with a non-trivial probability $\alpha < 1$. 
Assuming \text{LMA} is always enabled, all data views would come from IC-GAN generation. 
However, GAN is nutorious for mode collapse issue which lead a set of less diverse generated data than real dataset. 
The decreased diversity of training data would significantly hurt the performance of self-supervised representation learning, also evdienced by our experiment results (see Section~\ref{sec:prob_LMA} for analysis).
Hence, we mitigate this issue by occasionally applying \text{LMA}, which would make the source of training data a mixture of real data and generated one and thereby prevent reducing the diversity of training data.

\section{Experiments}
\subsection{Settings}
\label{sec:exp_setting}
\subsubsection{Datasets} 
Our method is evaluated on five datasets: CIFAR10, CIFAR100, STL10, ImageNet100, and ImageNet.
\textbf{CIFAR10} and \textbf{CIFAR100}~\cite{cifar_dataset2009Krizhevsky} are 32$\times$32-resolution image datasets with 10 and 100 classes, respectively. 
Both CIFAR10 and CIFAR100 are split into 50,000 images for training and  10,000 images for validation. 
\textbf{STL-10}~\cite{stl10_2011coates} and \textbf{ImageNet100}~\cite{cmc2020tian} are datasets derived from the ImageNet~\cite{imagenet2009deng}. 
STL-10 contains images at 96$\times$96 resolution of 10 classes, which are further split into training set with 5,000 labeled images plus  100,000 unlabeled images and test set with 8,000 labeled samples. 
ImageNet100 contains images of 100 classes, including a train split of 126,689 images and a validation split of 5,000 images.
\textbf{ImageNet}~\cite{imagenet2009deng} is the most popular large-scale image dataset of 1000 classes, which consists of 1,281,167 training images and 50,000 validation images.

\subsubsection{Training IC-GAN for LMA}
On CIFAR10, CIFAR100, and STL10, we first empoly SimSiam~\cite{simsiam2021chen} without \text{LMA} to learn a feature extractor which is later used for feature extraction of conditioning images.
Then we train an IC-GAN on train split for CIFAR10 and CIFAR100 with StyleGAN2 at $32\times32$ resolution as backbone, and on ``train+unlabel'' split of STL10 with StyleGAN2 at $128\times128$ resolution as backbone. 
For experiments on ImageNet100 and ImageNet, we utilize pre-trained IC-GAN that is publicly available\footnote{We use IC-GAN pretrained on ImageNet at $128\times128$ resolution with BigGAN as backbone: \url{https://dl.fbaipublicfiles.com/ic_gan/icgan_biggan_imagenet_res128.tar.gz}.}.
These pre-trained IC-GAN is repurposed for \text{LMA}.
When applying \text{LMA}, by default, we use $\alpha=0.3$ for CIFAR10, CIFAR100, STL10, and ImageNet100 and $\alpha=0.1$ on ImageNet.

\subsubsection{Integrating LMA into SSL}
We pre-train representation extractor on training set of each dataset with LMA-integrated SimSiam~\cite{simsiam2021chen} and MoCov2~\cite{mocov2_2020chen}.
For backbone feature extractor, we employ ResNet18~\cite{resnet2016he} on CIFAR10, CIFAR100, and STL10, where CIFAR variant of ResNet18~\cite{simsiam2021chen} is specifically utilized on CIFAR-10 and CIFAR-100.
ResNet50 is employed as backbone feature extractor on ImageNet100 and ImageNet. 
Other network details include projection (and prediction) heads follow the original practice of MoCov2 and SimSiam.
As in MoCov2~\cite{mocov2_2020chen}, the handcrafted augmentation includes random crop, color jittering, color discard, Gaussian blurring, and horizontal flip. Details are available in the appendix. 
SGD optimizer and cosine learning rate decay~\cite{cosinelr2016loshchilov} scheduler are used for training representation extractors. The actual learning rate is linearly scaled according to the ratio of batch size to 256, \ie $\texttt{base\_lr}\times\texttt{batch\_size}/256$~\cite{goyal2017accurate}. Detailed hyperparameters are available in the appendix.

\subsection{Main Results}

\begin{table}[t]
\centering
\tablestyle{4.5pt}{1.5}\scalebox{0.85}{\begin{tabular}{lcc|cccc}
    \shline
    Methods & HCA & LMA 
    & CIFAR10 & CIFAR100 & STL10 & IN100$^\dagger$ \\ 
    \hline
    \multirow{3}{*}{\tabincell{l}{\vspace{-6pt}SimSiam~\\\scalebox{0.7}{\cite{simsiam2021chen}}}}  & $\checkmark$ & 
    & 90.94 & 63.07 & 81.13 & 78.32 \\
    &  & $\checkmark$
    & 89.63 & 57.57 & 78.41 & 74.02 \\
    & $\checkmark$ & $\checkmark$
    & \textbf{92.46} & \textbf{65.70} & \textbf{81.92} & \textbf{82.94} \\  
    \hline
    \multirow{3}{*}{\tabincell{l}{\vspace{-6pt}MoCov2~\\\scalebox{0.7}{\cite{mocov2_2020chen}}}} & $\checkmark$ & 
    & 91.18 & 59.76 & 79.20 & 69.80 \\
    &  & $\checkmark$
    & 88.51 & 58.17 & 80.71 & 74.06 \\
    & $\checkmark$ & $\checkmark$
    & \textbf{92.02} & \textbf{64.89} & \textbf{82.72} & \textbf{80.80} \\  
    \shline 
\end{tabular}}
\caption{\textbf{Linear classification performance} on benchmarks at small and medium scales. ``HCA'' denotes the handcrafted augmentation. The top1 accuracy of the linear classifier atop pre-trained representations is reported. $^\dagger$: IC-GAN pre-trained on ImageNet-1K is employed for LMA on ImageNet100.}
\label{tab:gda_benchmark}
\end{table}

\begin{table}[t]
    \centering
    \tablestyle{7pt}{1.3}\scalebox{0.85}{\begin{tabular}{llc|c}
        \shline
        {Methods} & Sources & \# Epochs  & Top1 Acc \\ 
        \hline
        SimCLR & \citet{jahanian2021generative} & 20 & 43.90 \\ 
        on BigBiGAN Syn. & \citet{jahanian2021generative} & 20 & 35.69 \\
        + BigBiGAN-Aug & \citet{jahanian2021generative} & 20 & 42.58 \\
        \hline
        SimSiam & \citet{peng2022crafting} & 100 & 65.62 \\
        + ContrastiveCrop~ & \citet{peng2022crafting} & 100 & 65.95\\
        \hline
        MoCov2 $^\dagger$ & Ours & 100 & 62.48 \\
        + LMA & Ours & 100 & 63.97 \\
        SimSiam $^\dagger$ & Ours & 100 & 67.32 \\
        + LMA & Ours & 100 & 67.82 \\
        \shline 
    \end{tabular}}
    \caption{\textbf{Linear classification performance on ImageNet}.  $^\dagger$: Our reproduction results with $\alpha=0$.}
\label{tab:imagenet}
\end{table}

\subsubsection{Linear classification} 
Following common practice in SSL~\cite{simclr2020chen,cmc2020tian,moco2020he}, the quality of learned representations is evaluated with the performance of a trained linear classifier atop the representations.
Details are available in the appendix. 
Table~\ref{tab:gda_benchmark} compares the results of SimSiam~\cite{simsiam2021chen} and MoCov2~\cite{mocov2_2020chen} on small- and medium-scale benchmarks under the settings of (1) only using handcrafted augmentation (HCA), (2) only using LMA, and (3) using both HCA and LMA (see Algorithm~\ref{alg:gda_ssl}).
It can be seen that supplement of LMA improves (setting 3 v.s. 1) the top-1 accuracy of SimSiam with  1.52\%, 2.63\%, 0.79\%, and 4.62\% and MoCov2 with 0.84\%, 5.13\%, 3.52\%, and 11.00\% on CIFAR10, CIFAR100, STL10, and ImageNet100, respectively.
It is also noteworthy that by only using LMA the performance is not significantly decreased and sometimes surpass only using HCA: MoCov2 achieves 80.71\% with only LMA against 79.20\% on STL10, and 74.06\% agianst 69.80\% on ImageNet100.

Our method is further evaluated on the most popular large-scale dataset, ImageNet.
In particular, we pretrain a ResNet50 with SimSiam for 100 epochs, Table~\ref{tab:imagenet} presents the results of SimSiam and MoCov2 with LMA as well as other augmentation-related methods for reference. 
It shows that LMA can consistently bring clear improvement, with 0.50\% and 1.12\% increase on top-1 accuracy for SimSiam and MoCov2. 
Our method also significantly outperforms augmentation by traversing BigBiGAN latent space~\cite{jahanian2021generative}, presenting a more promising way to realizing GAN-based augmentation.

\subsubsection{Representation invariance} 
As LMA introduces additional nuisance variation such as object pose, viewpoint, lighting condition, \etc, we further evaluate if the learned representations gain stronger invariance to such variation.
To quantitatively measure the representation invariance, we follow~\citet{ericsson2021self} to extract representations and compute average pairwise cosine similarity for real-world images from datasets including Flickr1024~\cite{scharstein2014high_flickr}, COIL100~\cite{nene1996columbia_coil}, ALOI~\cite{geusebroek2005amsterdam_ALOI}, ALOT~\cite{burghouts2009material_ALOT}, ExposureErrors~\cite{afifi2021learning_ExposureErrors}, RealBlur~\cite{rim2020real_realblur} that are collected with controlled variation such as stereo, pose/scale, viewpoint, illumination, color temperature, exposure, and blurring.  
Fig.~\ref{fig:teaser_invariance} compares the representation invariance learned by MoCov2 with HCA, with LMA, and with HCA+LMA on ImageNet100.
Detailed numbers are available in the appendix. 
Results show that LMA excels at invariance to pose, viewpoint, exposure and illumination but compromises in stereo, blur, and color temperature agianst HCA. 
Supplementing LMA to HCA is able to bring gain more invariance without losing original invariance much.

\begin{table}[t]
    \centering
    \tablestyle{5pt}{1.3}\scalebox{0.95}{\begin{tabular}{lcccccc}
        \shline
        {Methods} & \tabincell{c}{\vspace{-5pt}IN-V2 \\(Top)} 
                  & \tabincell{c}{\vspace{-5pt}INv2 \\(Th0.7)} 
                  & \tabincell{c}{\vspace{-5pt}INv2 \\(Freq)} 
                  & IN-R & IN Sketch & IN-A \\ 
        \hline
        MoCov2    & 63.88 & 57.78 & 49.21 & 23.73 & 14.22 & \textbf{2.06} \\
        + LMA     & \textbf{66.61} & \textbf{59.40} & \textbf{50.65} & \textbf{25.22} &	\textbf{15.49} &	1.74 \\
        SimSiam   & 70.33 & 64.47 & 55.29 &	28.23 &	18.08 &	\textbf{2.64} \\
        + LMA     & \textbf{70.58} & \textbf{64.55} & \textbf{55.55} &	\textbf{29.41} & \textbf{18.48} & 2.58\\
        \shline 
    \end{tabular}}
    \caption{\textbf{Robustness evaluation}. Top-1 accuracy on ImageNet-like datasets with distribution shift.}
\label{tab:robustness}
\end{table}

\subsubsection{Robustness to distribution shift} 
We further test the robustness of learned representation. 
In particular, we use the ImageNet-testbed~\cite{taori2020measuring}
to test the feature extractor plus linear classifier head that are trained on ImageNet train split on several ImageNet-like datasets with distribution shift: ImageNet-V2 (IN-V2)~\cite{recht2019imagenetv2} 
including topimages (Top), threshold0.7 (Th0.7), and matched frequency (Freq) splits, ImageNet-R (IN-R)~\cite{hendrycks2021many_inr}, ImageNet Sketch (IN Sketch)~\cite{wang2019learning}, and ImageNet-A (IN-A)~\cite{hendrycks2021natural_ina}.
Results in Table~\ref{tab:robustness} show that with addition of LMA, the linear classifier can achieve higher performance on five out of six datasets with distribution shift, suggesting that the self-supervised representations strengthened by LMA gain stronger robustness.

\begin{figure}[t]
    \centering
    \includegraphics[width=0.92\linewidth]{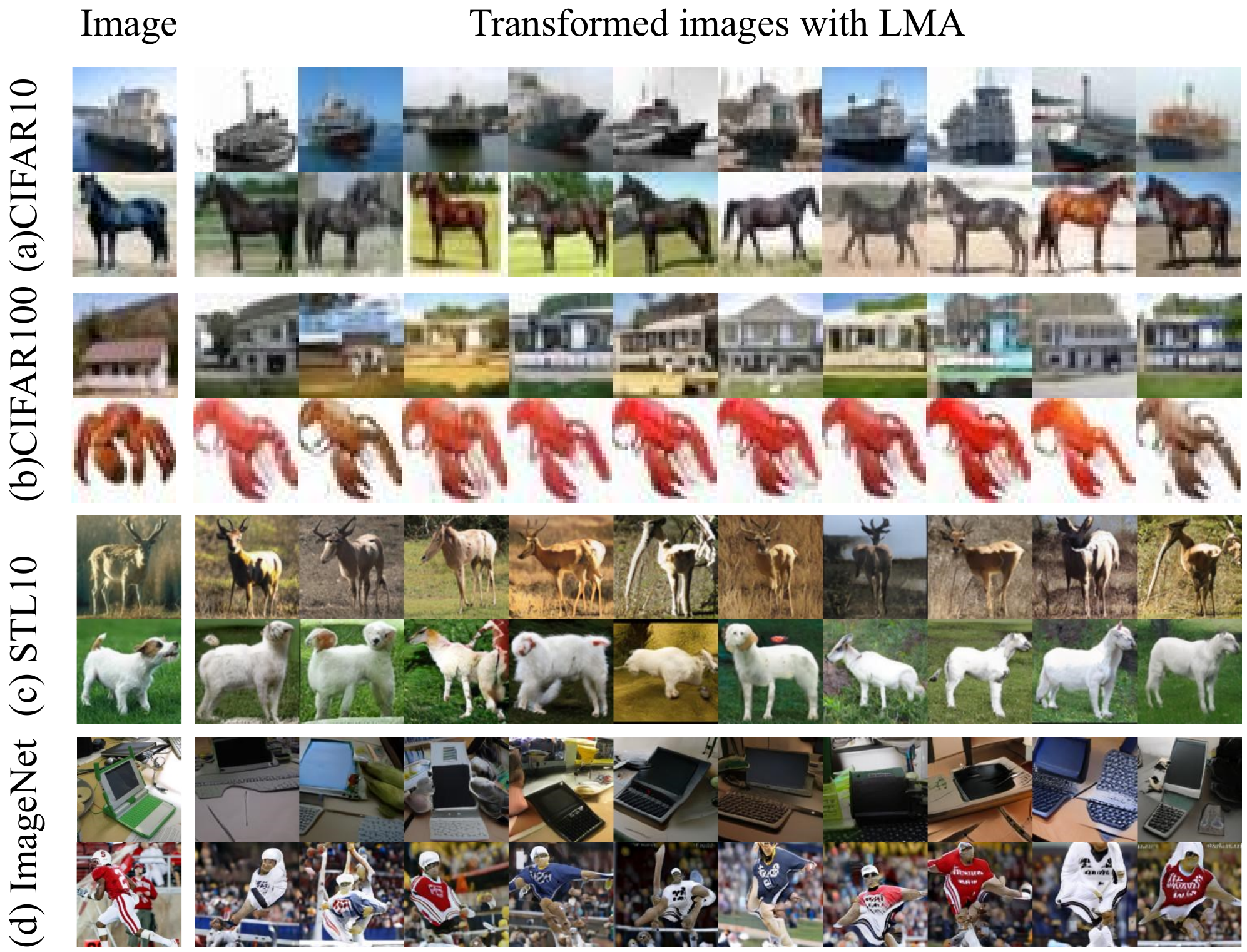}
    \caption{Visualization of LMA effects.}
    \label{fig:vis_lma}  
    \end{figure}

\begin{figure}[t]
\centering
\includegraphics[width=0.95\linewidth]{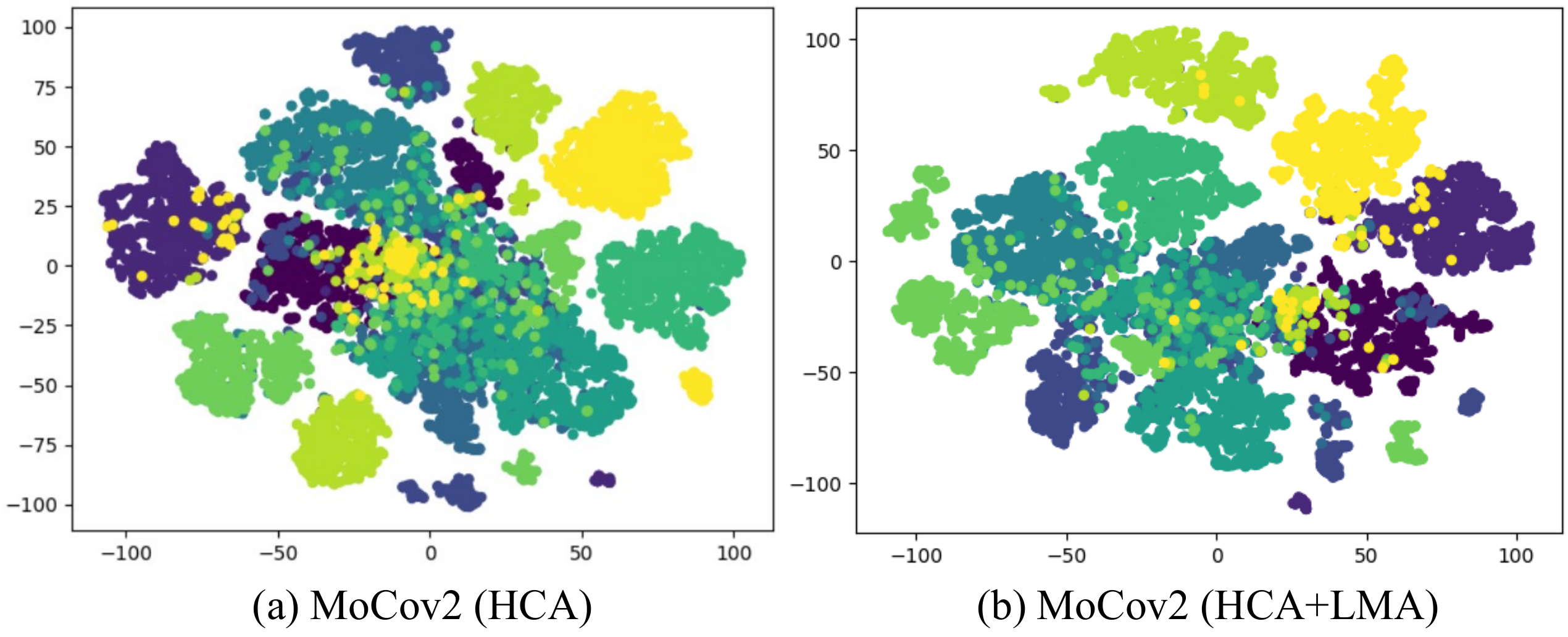}
\caption{Visualization of tSNE embedding of representations on CIFAR10 validation set.}
\label{fig:tsne}
\end{figure}

\subsubsection{Visualization} 
We provide visualization of LMA effects in Fig.~\ref{fig:vis_lma} and embedded representation distribution in Fig.~\ref{fig:tsne}.
It can be observed that LMA is generally able to preserve the semantic contents and introducing non-negligible variation.
With the help of LMA, the embedded representations are more discriminative for image classes.

\subsection{Analysis}

\subsubsection{Comparison to other transformation}

\begin{table}[t]\centering
    \tablestyle{4pt}{1.2}\begin{tabular}{l|cc|cc}
        \shline
        \multirow{2}{*}{LMA variants} & \multicolumn{2}{c|}{CIFAR10} & \multicolumn{2}{c}{CIFAR100} \\
        & Top1 Acc & Top5 Acc & Top1 Acc & Top5 Acc \\ 
        \hline
        w/o LMA 
        & 90.94 & 99.60 & 63.07 & 87.56 \\
        \hline
        $k$NN 
        & 89.56 & 99.52 & 63.14 & 87.84 \\
        StyleGAN2 
        & 91.67 & \textbf{99.80} & 63.81 & 88.75 \\
        * IC-GAN   & \textbf{92.46} & 99.73 & \textbf{65.70} & \textbf{89.92} \\
        \shline 
    \end{tabular}
    \caption{\textbf{Comparison of multiple LMA variants} on CIFAR10 and CIFAR100. The base SSL method is Simsiam. For $k$NN transformation, we use $k=20$ for CIFAR10 and $k=5$ for CIFAR100, the same as ones used for training IC-GAN. All the transformations are applied with probability 0.3. * denotes the default setting.}
\label{tab:other_aug}
\end{table}

In Table~\ref{tab:other_aug}, we compare our method to $k$NN transformation and traversing GAN.
In particular, $k$NN transformation approach uses as embedding network the same feature extractor as in pre-trained IC-GAN. 
It transforms a given image by replacing it with a random sample from its $k$ nearest neighbors (see Equ.~\ref{eq:knn_t}). 
Traversing GAN approach traverses the latent space (see Equ.~\ref{eq:trav_gan}) of a StyleGAN2~\cite{stylegan2-ada2020karras} generator that is pretrained on CIFA10 and CIFAR100. 
The perturbation is sampled from a Gaussian distribution with smaller std, \ie $\epsilon\sim N(0, 0.2)$.

According to the results, $k$NN transformation is not observed to consistently improve the performance of SimSiam and traversing GAN can bring slight improvement.
In contrast, our method clear outperforms other related transformation and contribute significant improvement to SimSiam.
As explained in the previous section, we attribute the success of LMA to its ability to generate infinite number of data of appropriate views.
To further study the effect of the number of views, we weaken LMA to only favor finite-view generation.
Concretely, we change the prior distribution of latent variables to a uniform distribution over a pre-sampled set.  
In this way, the \text{LMA} is restricted to create finite number of views. 
Table~\ref{tab:finite_LMA} shows that limiting the number of views significantly reduce the performance and this issue can be mitigated by increasing the number of views.
These results verify our conjection that the ability of creating infinite number of views is an critial ingredient that \text{LMA} contributes to the representation performance improvement.

\begin{table}[t]\centering
\tablestyle{5pt}{1.2}\scalebox{0.9}{\begin{tabular}{c|c|ccccc}
    \shline
     & {w/o LMA} & \multicolumn{5}{c}{w/ LMA} \\
    \hline
    {\# LMA views} & -- & 1 & 50 & 500 & 5000  & * Inf \\ 
    Top1 Acc   & 90.94 & 90.83 & 91.13 & 91.42 & 91.06 &  \textbf{92.46}  \\
    \shline 
\end{tabular}}
\caption{\textbf{Ablation w.r.t. number of views} on CIFAR10.
* indicates the default setting. ``Inf'' denotes infinite number.}
\label{tab:finite_LMA}
\end{table}

\subsubsection{Probability of applying LMA}\label{sec:prob_LMA}


\begin{figure}[t]
    \centering
    \includegraphics[width=0.49\linewidth]{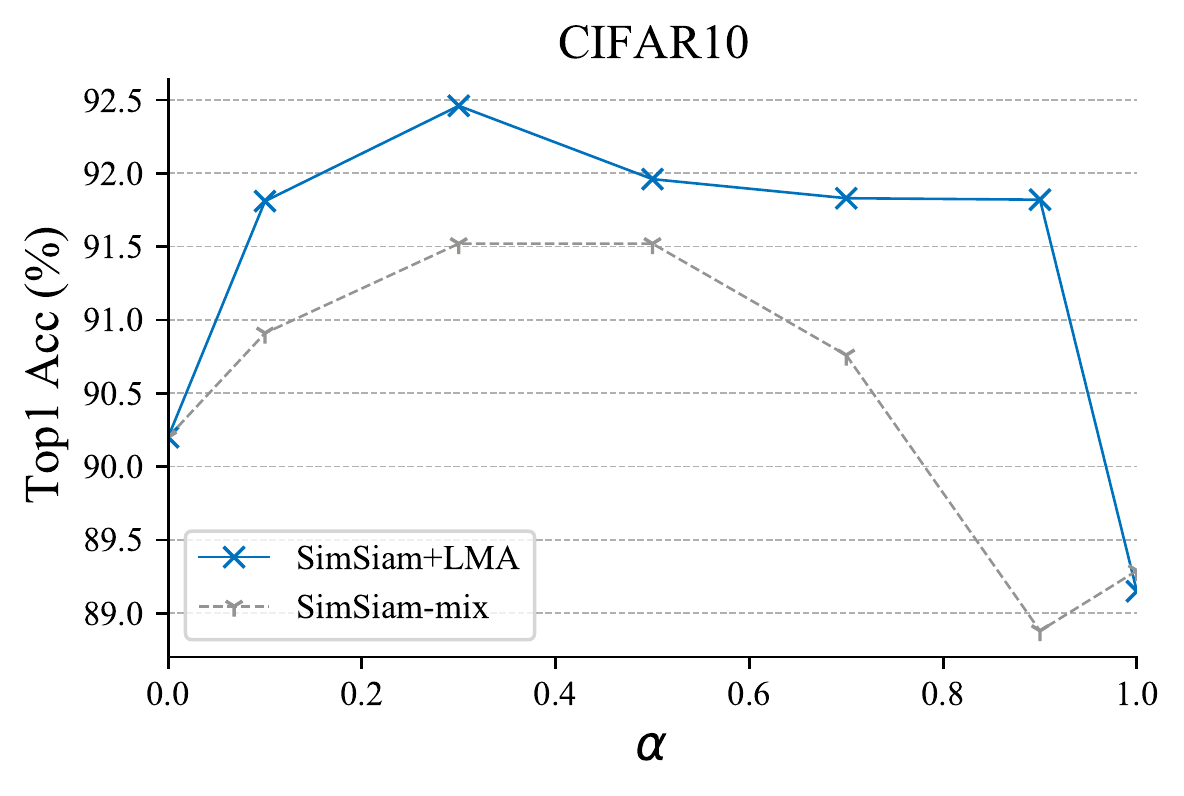}
    \includegraphics[width=0.49\linewidth]{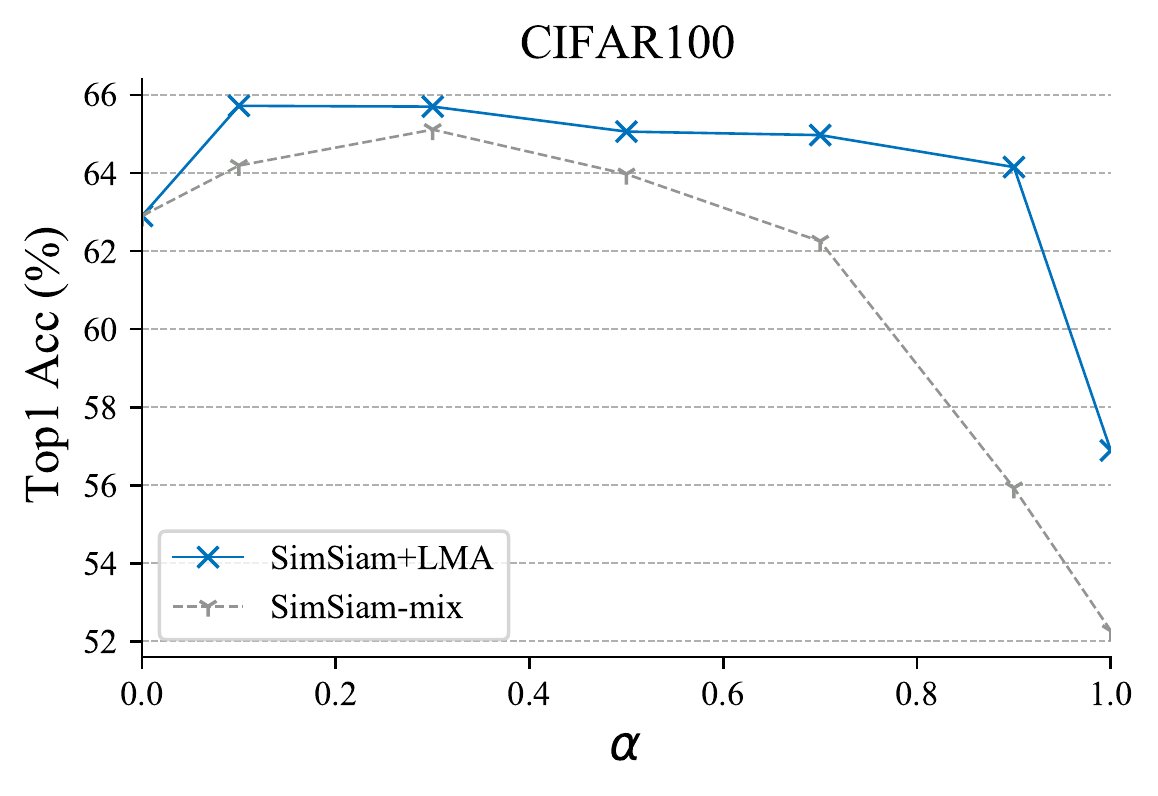}
    \caption{\textbf{Ablation w.r.t. $\alpha$.} The linear classification top1 accuracy on CIFAR10 and CIFAR100 with different probabilities of applying {LMA}.}
    \label{fig:ablation_p}
\end{figure}

As analyzed in Section~\ref{sec:ssl_LMA}, the introduction of {LMA} makes two changes to the training data: (1) the source of training images, \ie real data or generated data, and (2) multiview training data, \ie the view variation of the data.
Since LMA relies on generative model to create multiview data, it is impossible to administer the second ingredient without the first ingredent.  
To solely impose the first ingredient, we consider a reference method, ``SimSiam-mix'', where in a mini-batch of training data, training images are possible to be sampled from both real dataset and generator.
Similarly to Algorithm~\ref{alg:gda_ssl}, a hyperparameter $\alpha$ controls the possibility of sampling from generator.  
In this way, for both ``SimSiam-mix'' and ``SimSiam+LMA'', the soure of training data would be (1) real data when $\alpha=0$, (2) a mixture of real and generated data when  $0<\alpha<1$, and (3) generated data when $\alpha=1$.
Beyond that, ``SimSiam+LMA'' enjoys richer multiview data from LMA compared to ``SimSiam-mix''.

Fig.~\ref{fig:ablation_p} plots the performance of these two methods with respect to different $\alpha$.
It can be observed that the optimal $\alpha$ is around 0.3 for both CIFAR10 and CIFAR100.
Note that the performance when $\alpha=1$ is lower than performance when $\alpha=0$ for these two methods, suggesting that the pre-trained IC-GAN is unable to generate data of matching quality to real data.
Despite so, the performance of training on a mixture of data is higher than that on real data (see the performance of SimSiam-mix when $\alpha=0.3$ versus $\alpha=0$), suggesting that generated data does have some complementary effect to real data.
Finally, ``SimSiam+LMA'' clearly outperforms ``SimSiam-mix'', indicating that extra data variation from LMA do contribute to the improvement of representation learning. 

\subsubsection{The impact of pre-trained IC-GAN}


\begin{figure}[t]
\begin{floatrow}
    \capbtabbox[0.43\textwidth]{
    \centering
    \tablestyle{4pt}{1.2}\scalebox{0.85}{\begin{tabular}{lrc|c}
        \shline
        & $k$ & FID & {Top-1 Acc} \\
        \hline
        \multirow{3}{*}{C10} & 5    & 3.11 & 91.79 \\ 
        & * 20 & 2.84 & 92.46 \\ 
        & 50   & 2.77 & 92.51 \\
        \hline
        \multirow{3}{*}{C100} & 3   & 4.50 & 62.77 \\
        & * 5 & 4.41 & 65.70 \\
        & 10  & 4.39 & 66.50 \\
        \shline 
    \end{tabular}}
    }{
        \caption{Ablation w.r.t. $k$NN when pretraining IC-GAN. * denotes default setting. SimSiam+\text{LMA}}
        \label{tab:ablation_pretrain}
    }
    \ffigbox[0.45\textwidth]{
      \centering
      \includegraphics[width=0.95\linewidth]{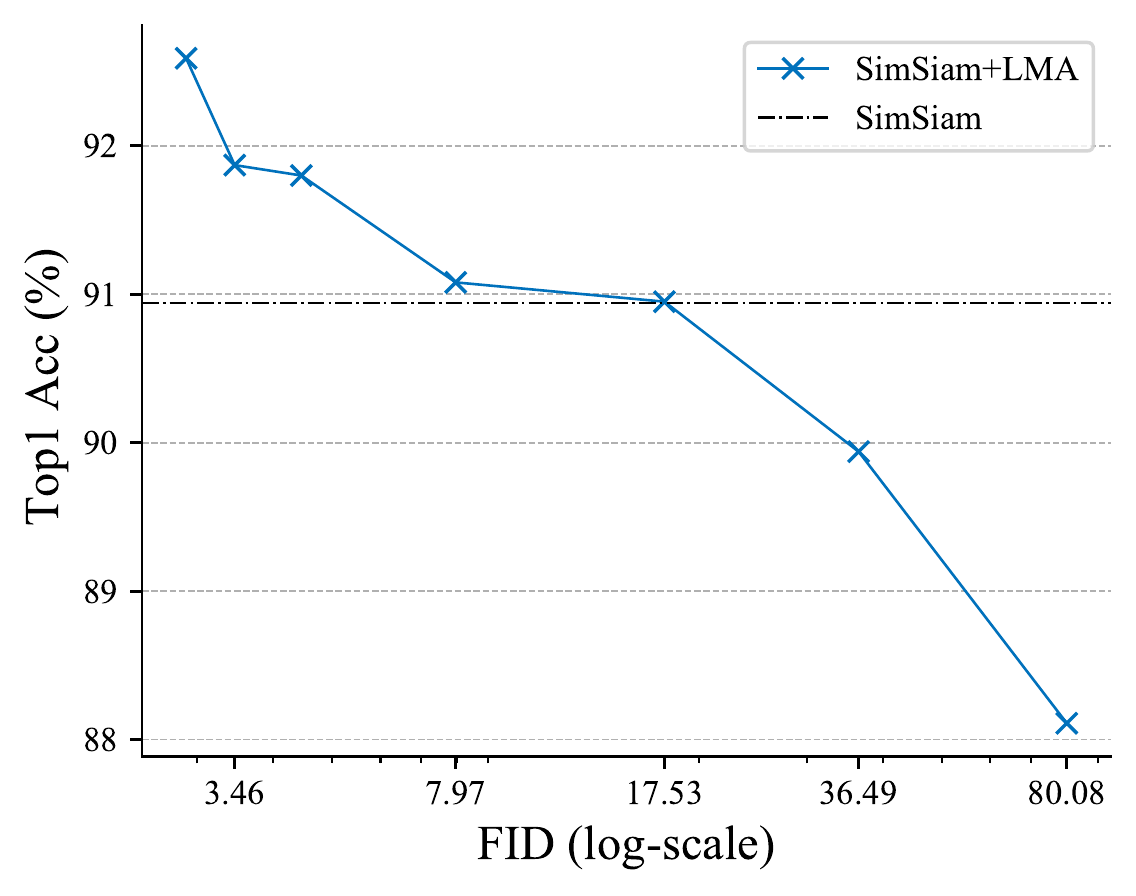}
    }{
      \caption{Ablation w.r.t. IC-GAN quality. The linear classification top1 accuracy of SimSiam+LMA versus FID of IC-GAN on CIFAR10.}
      \label{fig:ablation_icgan}
    }
\end{floatrow}
\end{figure}

Since \text{LMA} heavily relies on the pre-trained IC-GAN, we study how is the performance of our method related to the pre-trained IC-GAN.
We ablate the $k$ in $k$NN when pre-training IC-GAN and show the results in Table~\ref{tab:ablation_pretrain}.
Generally, if $k$ is increased, the IC-GAN can gain higher quality of generated data, suggested by lower FID and higher performance of ``SimSiam-mix''. 
It can be seen that with improved quality of pre-trained IC-GAN, the performance of our method can be further improved. 

Furthermore, we use various IC-GANs that are not well-trained for LMA to investigate the impact of the quality of IC-GAN on the performance our method.
In particular, we select the checkpoints of different FID that are saved during training process and run SimSiam with LMA ($\alpha=0.3$) on CIFAR10.  
Fig.~\ref{fig:ablation_icgan} plots the linear classification top-1 accuracy versus FID.
It shows that LMA can provides more improvement given a IC-GAN of higher quality.



\section{Conclusions}

In this paper, we construct local manifold augmentation (LMA) motivated by utilizing the rich data variation underlying the dataset. 
This is achieved by repurposing a pre-trained IC-GAN for data augmentation.
LMA is able to provide richer data variation that includes complicated geometrical and appearance change and able to improve SSL performance, representation invariance and representation robustness. 
Dedicated steeration to purify the useful data variation for representation learning is important to fine-grained recognition tasks and complicated and variable real-world scenarios, which is a open challenge left as future work. 

\section*{Acknowledgments}
This work was supported by the National Key R\&D Program of China under Grant 2018AAA0102801, National Natural Science Foundation of China under Grant 61620106005.


\bibliography{aaai23}




\appendix
\section{Implementation Details}
\subsection{Training IC-GAN}
Training IC-GAN requires a pre-trained feature extractor to embed images into feature vectors for nearest neighbor search and condition input.
On CIFAR10, CIFAR100, and STL10, we employ SimSiam~\cite{simsiam2021chen} with only handcrafted augmentation to learn such feature extractors.
Table~\ref{tab:origin_perfromance} summarizes the top-1 and top-5 accuracies of feature extractor as well as other hyperparameters for training IC-GANs. 
The generator with the lowest Fréchet Inception distance (FID) during training process is chosen and repurposed for LMA.

\begin{table}[t]\centering
	\newcommand{\wrapcell}[1]{\tabincell{c}{\vspace{-5pt}#1}}
	\newcommand{\tworow}[1]{\multirow{2}{*}{#1}}
	\tablestyle{3pt}{1.5}\scalebox{0.77}{\begin{tabular}{c|cc|ccccc}
			\shline
			\tworow{Dataset} & \multicolumn{2}{c|}{Feature extractor} 
			& \tworow{Split} & \tworow{Res.} & \tworow{Backbone} & \tworow{Cfg.} & {Duration} \\
			& Top1 Acc & Top5 Acc &  & &  &  & (kimg) \\ 
			\hline
			CIFAR10     & 90.94 & -   & \texttt{train} & 32 & StyleGAN2 & cifar & 100,000 \\
			CIFAR100    & 63.07 & 87.56 & \texttt{train} & 32 & StyleGAN2 & cifar & 100,000 \\
			STL10       & 81.13 & 98.88 & \wrapcell{\texttt{train}\\\texttt{+unlabel}} & 128 & StyleGAN2 & auto & 25,000\\
			\shline 
	\end{tabular}}
	\caption{Hyperparameters for training ICGAN on CIFAR10, CIFAR100, and STL10.}
	\label{tab:origin_perfromance}
\end{table}

\subsection{Handcrafted data Augmentation}
The construction of handcrafted augmentation follows the prevalent practice~\cite{mocov2_2020chen}, which is composed in the following sequence.
\begin{itemize}
	\item \texttt{RandomResizedCrop} that crops random patches with their area in [0.2,~1.0] and aspect ratio in [3/4,~4/3] from images  and resizes the patches into the input scale (see Table~\ref{tab:hyper_parameters_table} for specific input scale on each dataset).
	\item  \texttt{ColorJitter} that randomly scales the brightness, contrast, and saturation with factors in [0.6,~1.4], and the hue with factors in [-0.1,~0.1]. The ColorJitter is applied with a probability of 0.8.
	\item \texttt{RandomGrayscale} that randomly convert RGB images to the gray-scale ones. The RandomGrayscale is applied with a probability of 0.2.
	\item \texttt{GaussianBlur} that blurs images using Gaussian kernels with radius randomly sampled in [1,~2]. The GaussianBlur is applied with probability of 0.5 (disabled on CIFAR10 and CIFAR100).
	\item \texttt{RandomHorizontalFlip} that flips images horizontally with a probability of 0.5.
\end{itemize}

\subsection{SSL algorithm hyperparameters}

Table \ref{tab:hyper_parameters_table} summarizes the hyper-parameters for training Simsiam and MoCov2.

\subsection{Linear classification evaluation}
The linear classifier is trained using SGD with LARS~\cite{lars2017you} with base learning rate 0.1, momentum 0.9, weight decay 0., batch size 4096, and for 90 epochs.

\begin{table}[t]\centering
	\tablestyle{3pt}{1.5}\scalebox{0.9}{\begin{tabular}{c|c|ccccc}
			\shline
			Method & Dataset & Input scale & Backbone & Weight decay & Base lr \\ 
			\hline
			\multirow{5}{*}{SimSiam}
			& IN     & 224 & R50 & 0.001 & 0.05\\
			& IN100     & 96 & R50 & 0.001 & 0.05 \\
			& CIFAR10     & 32 & R18-C  &0.005& 0.03\\
			& CIFAR100    & 32 & R18-C &0.005& 0.03 \\
			& STL10       & 128 & R18 &0.005& 0.05 \\
			\shline 
			\multirow{4}{*}{MoCov2}
			& IN     & 224 & R50 & 0.001 & 0.03\\
			& IN100     & 96 & R50 & 0.001 & 0.03\\
			& CIFAR10     & 32 & R18-C &0.005 & 0.03\\
			& CIFAR100    & 32 & R18-C & 0.005& 0.03\\
			& STL10       & 128 & R18 & 0.005 &0.03 \\
			\shline 
	\end{tabular}}
	\caption{Hyper-parameters for SSL training on ImageNet, ImageNet100, CIFAR10, CIFAR100 and STL10. 
		``R50'', ``R18'', and ``R18-C'' represent ResNet50, ResNet18, and ResNet18 of CIFAR variant, respectively.}
	\label{tab:hyper_parameters_table}
\end{table}

\section{Visulization}
\begin{figure*}[t]
	\centering
	\includegraphics[width=0.9\linewidth]{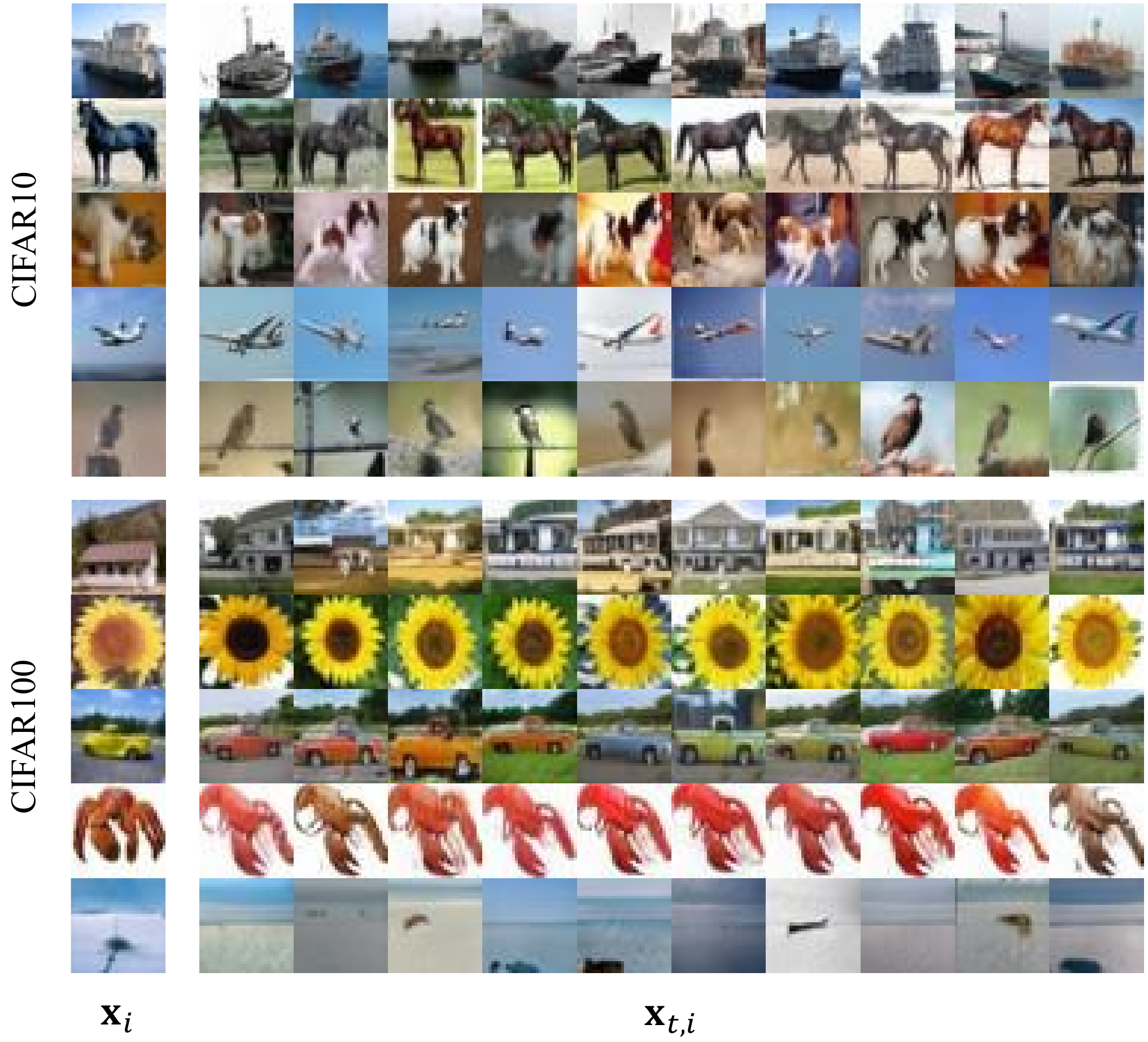}
	\caption{\textbf{Visualization} of LMA on CIFAR10 and CIFAR100.}
	\label{fig:vis_lma_cifar}
\end{figure*}

\begin{figure*}[t]
	\centering
	\includegraphics[width=0.9\linewidth]{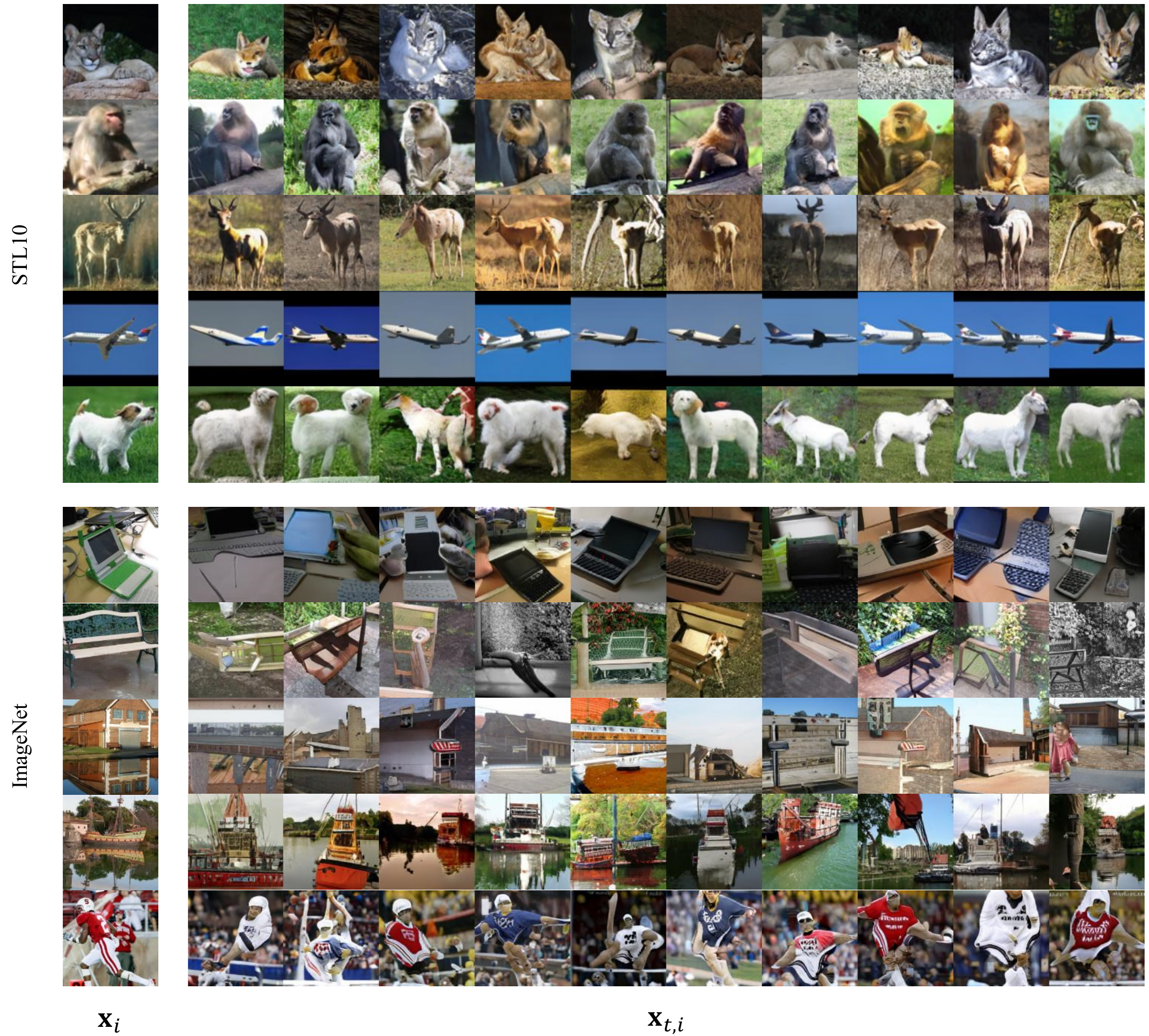}
	\caption{\textbf{Visualization} of LMA on STL10 and ImageNet.}
	\label{fig:vis_lma_stl_in}
\end{figure*}

Please see Fig.~\ref{fig:vis_lma_cifar} and Fig.~\ref{fig:vis_lma_stl_in} for more visualization of LMA effects.

\section{More Evaluation Results}

\begin{table}[t]
	\centering
	\tablestyle{10pt}{1.25}\begin{tabular}{l|ccc}
		\shline
		\multirow{2}{*}{Methods}
		& \multicolumn{3}{c}{PASCAL VOC} \\ 
		& AP$_{50}$ & AP & AP$_{75}$ \\
		\hline
		MoCov2 & 79.42 & 53.64 & 58.92 \\
		+ LMA & \textbf{79.95} & \textbf{53.78} & \textbf{59.19} \\  
		\shline 
	\end{tabular}
	\caption{\textbf{Transfer learning evaluation}. Backbone encoder is pre-trained on ImageNet100, and IC-GAN pre-trained on ImageNet-1K is employed for LMA. }
	\label{tab:gda_transfer}
\end{table}

\begin{table}[t]\centering
	\tablestyle{4pt}{1.2}\begin{tabular}{l|cc|cc}
		\shline
		\multirow{2}{*}{Methods}
		& \multicolumn{2}{c|}{1\% label} & \multicolumn{2}{c}{10\% label} \\ 
		& Top-1 Acc & Top-5 Acc & Top-1 Acc & Top-5 Acc \\
		\hline
		MoCov2 & 33.80 & 53.90 & 72.60 & 92.40 \\
		+ LMA & 46.50 & 66.80 & 78.10 & 94.80 \\  
		\shline 
	\end{tabular}
	\caption{\textbf{Semi-supervised evaluation} on ImageNet100. Backbone encoder is pre-trained on ImageNet100}
	\label{tab:gda_semi}
\end{table}

\begin{table*}[t]
	\centering
	\tablestyle{4pt}{1.2}\begin{tabular}{l|cccccccccc}
		\shline
		Variation  &  Stereo & Pose/Scale & \multicolumn{2}{c}{Viewpoint} & \multicolumn{2}{c}{Illumination} & \multicolumn{2}{c}{Temperature} & Exposure & Blur \\ 
		Dataset    & Flickr1024 & COIL100 & ALOI & ALOT & ALOI & ALOT & ALOI & ALOT & ExposureErrors & RealBlur \\ 
		\hline
		\multicolumn{11}{c}{Cosine similarity ($\uparrow$)} \\
		\hline
		MoCov2 w/ HCA & 0.94 & 0.75 & 0.79 & 0.68 & 0.84 & 0.70 & 0.98 & 0.95 & 0.86 & 0.92 \\
		MoCov2 w/ LMA & 0.92 & 0.81 & 0.80 & 0.72 & 0.86 & 0.81 & 0.97 & 0.94 & 0.88 & 0.84 \\
		MoCov2 w/ HCA + LMA  & 0.93 & 0.79 & 0.81 & 0.71 & 0.86 & 0.76 & 0.98 & 0.95 & 0.90 & 0.89 \\
		\hline
		\multicolumn{11}{c}{Mahalanobis distance ($\downarrow$)} \\
		\hline
		MoCov2 w/ HCA & 17.24 & 28.10 & 19.87 & 46.82 & 18.86 & 48.80 & 4.72 & 24.38 & 20.09 & 15.31 \\
		MoCov2 w/ LMA  & 26.41 & 28.67 & 23.31 & 49.93 & 20.36 & 44.28 & 9.15 & 30.18 & 23.87 & 31.39 \\
		MoCov2 w/ HCA + LMA & 18.34 & 25.97 & 18.14 & 41.99 & 16.51 & 40.93 & 6.12 & 22.85 & 16.29 & 19.88 \\
		\shline 
	\end{tabular}
	\caption{Real-world transformation invariance is measured with cosine similarity ($\uparrow$) and Mahalanobis distance ($\downarrow$)following~\citet{ericsson2021self}. The feature extractor is pre-trained on ImageNet100 using MoCov2.}
	\label{tab:invariance}
\end{table*}

Besides this linear classification evaluation, we also conduct semi-supervised learning following~\cite{simsiam2021chen} to evaluate the representation quality and transfer learning following~\cite{moco2020he} to evaluate the transferability of the learned representations. 
Additional details about representation invariance evaluation are also appended.

\subsection{Transfer learning}
Following~\cite{moco2020he}, the transferability of learned representations is evaluated with object detection task on PASCAL VOC. 
In particular, we pre-train ResNet50 on ImageNet100, initialize the backbone of R50-C4 in Faster R-CNN with the pre-trained one, train Faster R-CNN on the VOC \texttt{trainval2007+2012} split, and report its performance on the VOC \texttt{test2007} split.
All network layers are trainable and finetuned during training. 
Tab.~\ref{tab:gda_transfer} presents the results, showing that LMA provides marginal improvements on downstream tasks. 

\subsection{Semi-supervised learning}
Similarly to~\cite{simclr2020chen}, after pre-training representations on ImageNet100, we leverage a small subset of the available labels in the ImageNet100 \texttt{train} split to finetune a classification network.
Table~\ref{tab:gda_semi} reports the top-1 and top-5 accuracy on the ImageNet100 \texttt{val} split.
The results show that large improvement can be obtained with the help of LMA, 12.70\%/12.90\% top-1/top-5 accuracy improvement when 1\% labels are used and 7.50\%/2.40\% top-1/top-5 accuracy improvement when 10\% labels are used.

\subsection{Representation invariance details}
We follow~\citet{ericsson2021self} to evaluate the invariance of representations with respect to various real-world transformation. 
Table~\ref{tab:invariance} presents the detailed numbers. 
We use the evaluation results of cosine similarity to plot the radar chart in the Fig.1b in the main text. 

\end{document}